\documentclass{article}
\usepackage{spconf,amsmath,graphicx}
\usepackage{cite}
\usepackage{hyperref}
\usepackage{amsmath,amssymb,amsfonts}
\usepackage{algorithmic}
\usepackage{graphicx}
\usepackage{stfloats}
\usepackage{textcomp}
\usepackage{xcolor}
\usepackage{booktabs}
\usepackage{enumitem}
\setlist{nosep, leftmargin=14pt}

\usepackage{mwe} 

\title{SelfMedHPM: Self Pre-training With Hard Patches Mining Masked Autoencoders For Medical Image Segmentation}
\name{Yunhao Lv$^1$ \qquad Lingyu Chen$^1$ \qquad Jian Wang$^2$ \qquad Yangxi Li$^{2\star}$ \qquad Fang Chen$^{2\star}$}
\address{$^{1}$ Department of Computer Science and Engineering, Nanjing University of Aeronautics and \\Astronautics, Nanjing, China\\
$^{2}$ School of Biomedical Engineering and the Institute of Medical Robotics, Shanghai\\
JiaoTong University, Shanghai, 200240, China
}
\begin{document}
%
\maketitle

\begin{abstract}
In recent years, deep learning methods such as convolutional neural network (CNN) and transformers have made significant progress in CT multi-organ segmentation. However, CT multi-organ segmentation methods based on masked image modeling (MIM) are very limited. There are already methods using MAE for CT multi-organ segmentation task, we believe that the existing methods do not identify the most difficult areas to reconstruct. To this end, we propose a MIM self-training framework with hard patches mining masked autoencoders for CT multi-organ segmentation tasks (selfMedHPM). The method performs ViT self-pretraining on the training set of the target data and introduces an auxiliary loss predictor, which first predicts the patch loss and determines the location of the next mask. SelfMedHPM implementation is better than various competitive methods in abdominal CT multi-organ segmentation and body CT multi-organ segmentation. We have validated the performance of our method on the Multi Atlas Labeling Beyond The Cranial Vault (BTCV) dataset for abdomen mult-organ segmentation and the SinoMed Whole Body (SMWB) dataset  for body multi-organ segmentation tasks. 

\end{abstract}
\section{Introduction}

Masked Autoencoder (MAE)\cite{1} has recently been shown to
be effective in pre-training Vision Transformers (ViT)\cite{2} for
medical image analysis\cite{3}.  \textbf{However, the MAE pre-training approach in medical image analysis does not determine where is hard to reconstruct, but we think learning to produce the patches which are hard to reconstruct is also crucial.} By learning to generate patches that are difficult to reconstruct, the model can be forced to have a more comprehensive understanding of the medical image content, resulting in more desirable tasks to guide themselves.
    
In this paper, we propose a self pre-training method based on Hard Patches Mining (HPM)\cite{4} for medical image segmentation tasks.
Specifically, given an input medical image, instead of randomly generating a binary mask, we first let the model as a teacher to generate a demanding mask and then train the model as a student to predict the mask patch. In this way, the model is urged to learn where it is worth being masked, and how to reconstruct it at the same time. 
Then, the question becomes how to design the auxiliary task, to make the model aware of where the hard patches are.
We think that those discriminative parts of a medical image (e.g., organs) are usually hard to reconstruct, resulting in larger losses. 
Therefore, by simply urging the model to predict reconstruction loss for each patch, and then masking those patches with higher predicted losses, we can obtain a more formidable MIM task. 
To achieve this, we introduce an auxiliary loss predictor in the pre-training, predicting patch-wise losses first and deciding where to mask next based on its outputs. 
We describe the segmentation pipeline based on HPM self-training as follows. First, we apply HPM pre-training on the train-set as the downstream segmentation task.
Next, the pre-trained ViT
weights of student model are transferred to initialize the segmentation encoder.
Then the whole segmentation network, e.g., UNETR\cite{5}, is finetuned to segment.

A comparative analysis 
shows that the proposed method achieves 
state-of-the-art performances. 
SelfMedHPM achieves 90.9\% and 85.8\%
best DSC on SMWB and BTCV\cite{6}, outperforming 
selfMedMAE by +2.5\% and +2.3\%, respectively.
\section{METHODOLOGY}
\label{sec:format}

We first give an overview of our proposed selfMedHPM.
Then, the two objectives, i.e., reconstruction loss and patches loss are introduced in
\ref{sec:1} and \ref{sec:2}, respectively.
Next, in \ref{sec:3}, the
easy-to-hard mask generation manner is described.
\begin{figure*}[t]
\centering
\vspace{-0.8cm}
\setlength{\belowdisplayskip}{3pt}
\includegraphics[width=\textwidth]{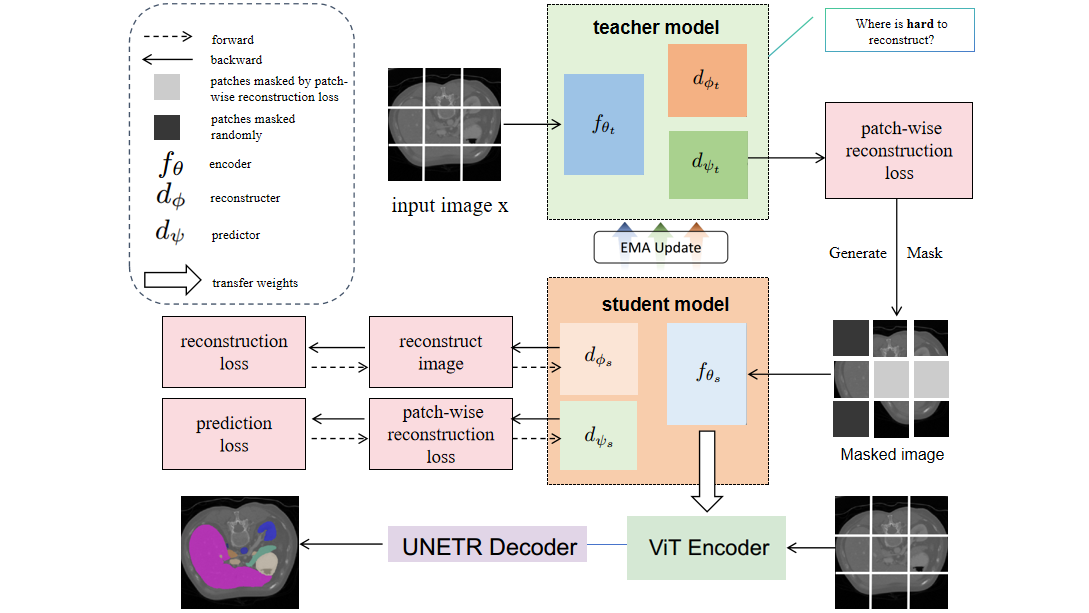}
\caption{\textbf{Illustration of our proposed selfMedHPM}, containing a student network and a teacher network, where the teacher is updated by the student in an exponential moving average (EMA) manner.} 
\label{fig1}
\end{figure*}
While traditional MIM pre-training solutions for medical image\cite{3,7} can be seen as training a student to solve a given problem, we consider it essential to make models stand in the teacher's position and generate a challenge pretext task. To achieve this, we introduce
an auxiliary decoder to predict the reconstruction loss of
each masked patch, instead of only the reconstruction loss of the entire image.
The student ($ f_{\theta _s},d_{\phi _s},d_{\psi _s}$) and the teacher ($ f_{\theta _\textit{t}},d_{\phi _\textit{t}},d_{\psi _\textit{t}}$)  have the same network architecture.
$ f_{\theta},d_{\phi},d_{\psi}  $ are encoder, image
reconstructor, and difficulty predictor, and the subscript t stands for teacher and s stands for student.
At each training iteration, an input 3D image $ I\in\mathbb{R}^{H\times W\times D\times C} $ is reshaped into a sequence of 2D patches $ x\in\mathbb{R}^{N\times (P^2C)} $.
$ (H,W,D)$ is the resolution of the original image, $C $ is the number of channels, $ P $ is the patch size, and $N = HW/P^2$ hence. Then, $x $ is fed into the teacher to get
patch-wise reconstruction loss $ \hat{L}^t$.
Based on $ \hat{L}^t$ and the training status, a binary mask $ M$ is generated under an easy-to-hard manner introduced  in
~\ref{sec:3}, the student is trained based on two objectives, i.e.,
reconstruction loss and prediction loss (~\ref{sec:2}).

After completing the self pre-training , we proceed
to fine-tune the model for the downstream segmentation task.
UNETR employs a U-Net-like\cite{8} design where encoder features
at various resolutions are skip-connected with the decoder.
We utilize the pre-trained ViT encoder of the student and append a
randomly-initialized convolutional decoder to UNETR.
\subsection{Reconstruction Loss}
\label{sec:1}
The reconstruction loss of our image reconstructor, $ L_{rec}$ remained consistent with selfMedMAE\cite{3}, i.e., mean squared error. Instead of reconstructing the complete image/volume, i.e., both visible and masked
patches, we only predicts the voxel values of the
masked patches, and normalized voxel values within each patch are reconstruction targets.

\subsection{Prediction Loss}
\label{sec:2}
In addition to designing mask strategies through prior knowledge, we believe that the ability to produce patches that are hard to reconstruct is also crucial for MIM pre-training.
Intuitively, we consider patches with high reconstruction loss as hard patches, which implicitly indicate the most
discriminative parts of an image.
To this end, we employ an extra difficulty predictor (i.e., $ d_\psi$) to mine hard patches during training. About how to design difficulty predictors, given a sequence of patch-wise reconstruction loss
$ L_{rec} \in\mathbb{R}^N$ of a image, the patchwise difficulty of the reconstruction task can be measured by $ argsort(L_{rec})$ . However, as the $argsort() $ operation
is non-differentiable.
Therefore, for each pair of patches $ (i; j)$, where $i; j = 1; 2;...; N$ and
$i\neq j$, we can implicitly learn $ argsort(L_{rec})$ by predicting
the relative relation of $ L_{rec}(i)$ and $ L_{rec}(j)$, i.e., which one
is larger. The objective is defined as follows:
\begin{align}
L_{pred}=
-\sum_{i=1}^{N} \sum_{j=1,j\neq i}^N \mathbb{I}_{ij}^{+} log(\sigma(\hat{L}_i^s - \hat{L}_j^s)-
    \notag
    \\ \sum_{i=1}^{N} \sum_{j=1,j\neq i}^N \mathbb{I}_{ij}^{-} log(1- \sigma(\hat{L}_i^s - \hat{L}_j^s)
\end{align}
where $ \hat{L}^s$ represents the patch-wise reconstruction loss from the student, and $i,j = 1,2,...,N$ are patch indexes.
\begin{figure}[t]
\includegraphics[width=\linewidth]{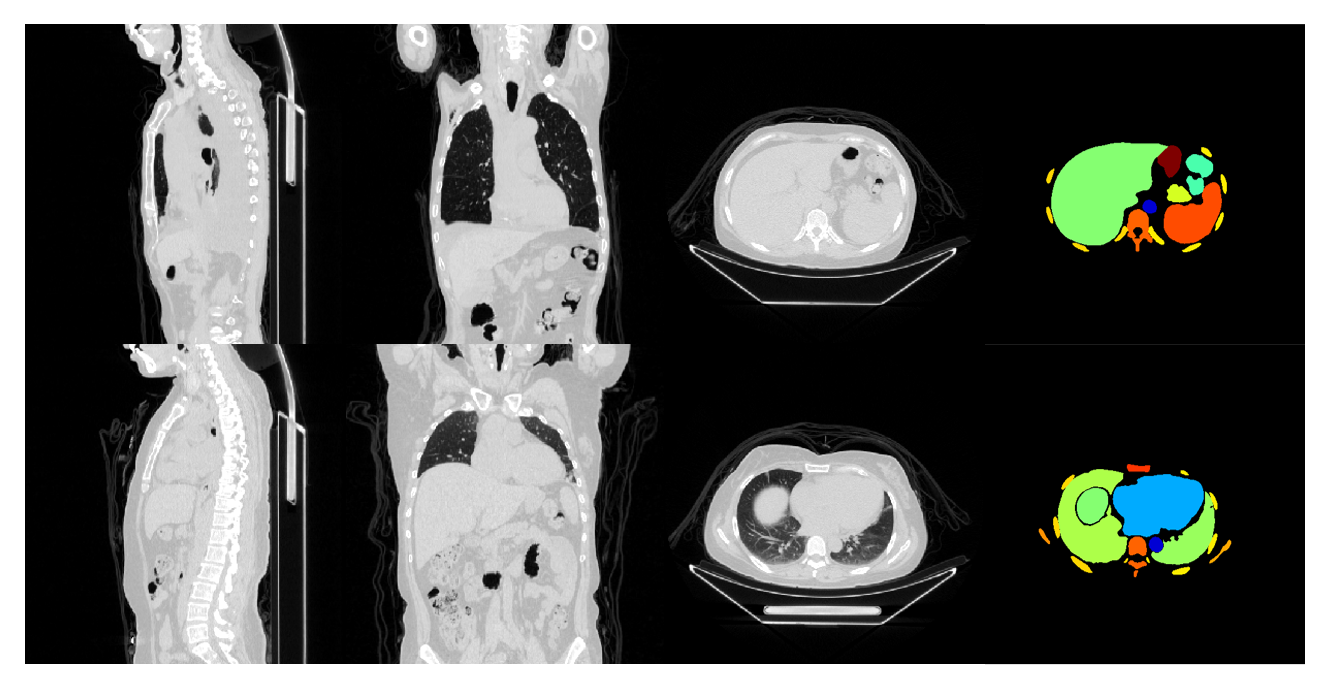}
\caption{Scans in the SMWB are carefully annotated with 41
organs. Each row from left to right is the scanning plane image, coronal plane image, axial plane image, and axial plane labels.}
\label{fig2}
\end{figure}
\begin{table*}[htbp]
\centering
\caption{Abdomen Multi-organ Segmentation on BTCV. }
\label{tab1}
\resizebox{\linewidth}{!}{
\begin{tabular}{@{\extracolsep{\fill}}cccccccccc}
\toprule
Framework &Avg DSC$\uparrow$/HD95$\downarrow$&Aorta&Gallbladder&Kidney(L)&Kidney(R)&Liver&Pancreas &Spleen &Stomach  \\
\midrule
	U-Net(R50)\cite{8}  &74.68/36.87 &84.18 &62.84 &79.19 &71.29 &93.35 &48.23 &84.41 &73.92\\
	AttnUNet(R50)\cite{10} &75.57/36.97 &55.92 &63.91 &79.20 &72.71 &93.56 &49.37 &87.19 &74.95\\ 
	TransUNet\cite{9}  &77.48/31.69 &87.23 &63.13 &81.87 &77.02 &94.08 &55.86 &85.08 &75.62                \\
	DSTUNet\cite{11}  &82.44/17.83 &88.16 &67.40 &87.46 &81.90 &94.17 &66.16 &92.13 &\textbf{82.10}\\
        UNETR\cite{5} &78.83/25.59 &85.46 &70.88 &83.03 &82.02 &95.83 &50.99 &88.26 &72.74\\
        UNETR+ImageNet  &79.67/24.28 &86.07 &74.29 &82.44 &81.65 &95.84 &58.08 &87.74 &69.98                  \\
        UNETR+MAE   &83.52/10.24 &88.92 &75.25 &86.37 &84.00 &95.95 &65.02 &90.56 &80.89\\
        UNETR+HPM   &\textbf{85.78/6.45} &\textbf{91.67} &\textbf{76.39} &\textbf{88.03} &\textbf{88.16} &\textbf{96.52} &\textbf{71.67} &\textbf{92.53} &\textbf{81.29}\\
\bottomrule
\end{tabular}}
\end{table*}
$ \sigma ()$ indicates sigmoid function. $ \mathbb{I}_{ij}^{+}$ and $ \mathbb{I}_{ij}^{-}$, are two indicators, representing the relative relationship of ground-truth reconstruction losses,
i.e., $ L_{rec}$, between patch $i$ and patch $j$. $ \mathbb{I}_{ij}^{+} = 1 $  if $ L_{rec}(i)>L_{rec}(j)$ and both patch $i$ and $j$ are masked else 0. $ \mathbb{I}_{ij}^{-} = 1 $  if $ L_{rec}(i)<L_{rec}(j)$ and both patch i and j are masked else 0.
\subsection{Easy-to-Hard Mask Generation}
\label{sec:3}
 In the early training stages, the learned feature
representations are not ready for reconstruction but are
overwhelmed by the rich texture, which means large
reconstruction loss may not be equivalent to discriminative.
To this end, we propose an easy-to-hard mask generation
manner, providing some reasonable hints that guide the
model to reconstruct masked hard patches step by step.
For each training epoch $ t$, $ \alpha_t$ of the
mask patches are generated by $ \hat{L}^t$ , and  the remaining $ 1-\alpha_t$ are randomly selected. Specifically, $ \alpha_t = \alpha_0 + \frac{t}{T}(\alpha_t - \alpha_0)$
, where $T$ is the total training epochs, and $ \alpha_0,\alpha_t \in [0,1]$ are
two tunable hyper-parameters. We filter $ \alpha_trN$ patches with
the highest $ \hat{L}^t$ to be masked, and the remaining $ (1-\alpha_t)rN$
patches are randomly masked, $ r$ is the mask ratio, typically $75$ percent. The proportion $ \alpha_t$ gradually increases from $ \alpha_0$ to $ \alpha_t$ in a linear manner, contributing to an easy-to-hard training procedure.
\begin{figure}[htbp]
\includegraphics[width=\linewidth]{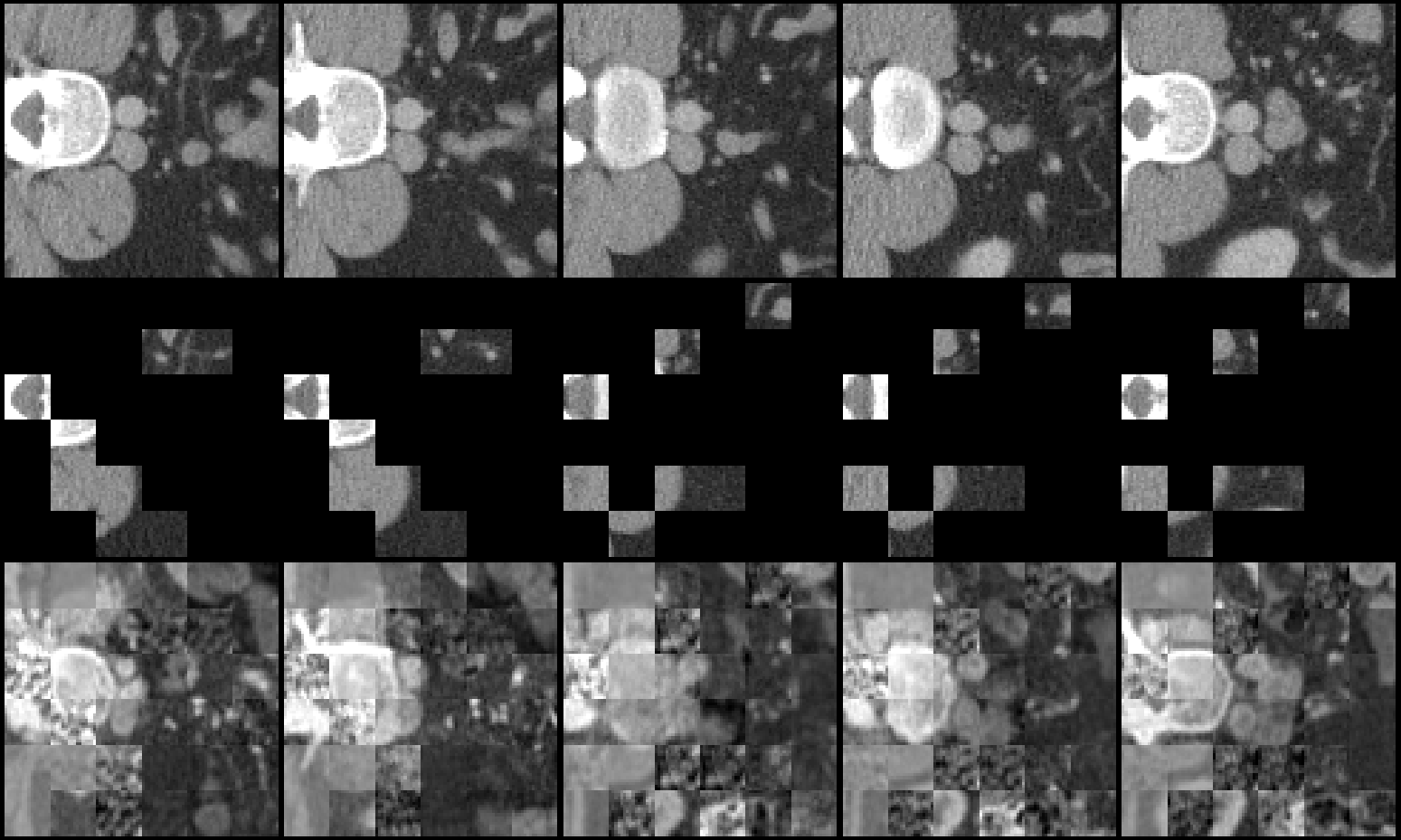}
\caption{\textbf{Reconstruction results of BTCV}. First row: Original image.
Second row: Masked image where masked regions are colored with black. Third row: Reconstructed images from
unmasked patches. Each column shows the slices of different depths.}
\label{fig3}
\end{figure}
\vspace{-0.7cm}
\section{EXPERIMENTS AND RESULTS}
\label{sec:pagestyle}
\subsection{Datasets and Implementation Details}
\textbf{Abdomen Multi-organ Segmentation on BTCV.} 
BTCV dataset consists of 30 cases of 3D abdominal multi-organ images and
each 3D image has 13 organ segmentation targets.
We report the average Dice similarity coefficient (DSC) and
95\% Hausdorff Distance (HD) on 8 abdominal organs to align with \cite{9} for ease of comparison. \\
\textbf{Body Multi-organ Segmentation on SMWB.} The SMWB included whole-body CT scans of 52 subjects, of which 41 organs were annotated by translation under physician supervision. Each CT scan included 493 slices, $512\times512$ pixels, with a voxel spatial resolution of (0.9766 × 0.9766 × 1.867 $ {mm}^3$). All data are resampled to the same space (3.5, 2.0, 2.0). We divided the 52 cases into 46 training sets and 6 test sets.
We reported the average DSC and the 95\% HD on eight representative organs of the chest and abdomen (heart, intestine, left lung, right lung, liver, pancreas, spleen, ventricle).
An example of image and annotation from the SMWB
dataset is shown in Fig. \ref{fig2}.\\
\begin{figure}[t]
\includegraphics[width=\linewidth]{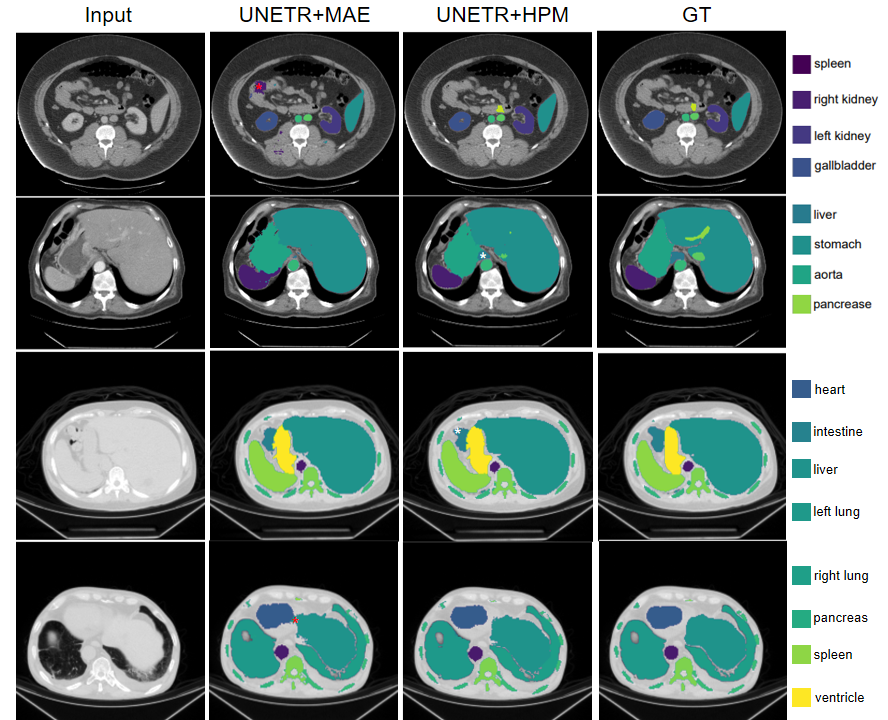}
\caption{\textbf{Qualitative Results of Segmentation. }Results for BTCV are shown in the first two rows. Results for SMWB are shown in the last two rows. In the first and fourth row, note that there is no false positive segmentation (red asterisk). In the second and third rows, the segmentation created by the MAE pre-trained UNETR method (white asterisk) is incomplete compared to the HPM pre-trained UNETR.}
\label{fig4}
\end{figure}
\begin{table*}[htbp]
\centering
\caption{Body Multi-organ Segmentation on SMWB. }
\label{tab2}
\resizebox{\linewidth}{!}{
\begin{tabular}{@{\extracolsep{\fill}}cccccccccc}
\toprule
Framework &Avg DSC$\uparrow$/HD95$\downarrow$&Heart&Intestine&Lung(L)&Lung(R)&Liver&Pancreas &Spleen &Ventricle  \\
\midrule
	U-Net(R50) &81.85/8.97 &87.45 & 81.88 & 92.34 & 92.11 & 90.75 & 57.12 & 86.03 & 75.10\\
	TransUNet  &83.37/7.44  &87.53 & 82.12 & 92.56 & 93.73 & 91.61 & 57.73 & 86.21 & 75.45               \\
        UNETR &85.88/7.24 &90.78&84.92&95.31&95.76&94.09&59.84&88.42&77.93\\
        UNETR+ImageNet  &86.44/6.84 &91.23&85.37&95.84&96.18&94.56&60.72&88.95&78.67                 \\
        UNETR+MAE   &88.46/3.53 &93.45 &87.50 &97.00 &97.20 &96.02 &64.08 &91.05 &81.36\\
        UNETR+HPM   &\textbf{90.93/2.65} &\textbf{94.60} &\textbf{90.06} &\textbf{97.87} &\textbf{98.10} &\textbf{96.65} &\textbf{72.51} &\textbf{93.19} &\textbf{84.42}\\
\bottomrule
\end{tabular}}
\end{table*}
\textbf{Implementation Details.}
Our experiments were implemented in PyTorch\cite{12} and MONAI\cite{13} and were conducted on a computer equipped with a NVIDIA RTX4070 GPU with 16GB of RAM. We use ViT-B/16 as the backbone and AdamW
as the optimizer in all the experiments. The patch size is 16 × 16 × 16.
We clip the raw values, re-scale the
range within [0,1], randomly flip and crop a 96 × 96 × 96 volume as the input and use a batch size of 4 for each dataset.
During pre-training, the initial learning rate (lr) is 1.5e-4
and weight decay is 0.05. Lr decays to zero following a cosine schedule with warm-ups.
During fine-tuning,  we adopt layer-wise learning rate decay
(layer decay ratio: 0.75) to stablize the ViT training and a
random DropPath with a 10\% probability. The learning rate is 8e-4. Learning rate
during fine-tuning also follows a cosine decay schedule.
\vspace{-0.35cm}
\subsection{Results}
\textbf{Reconstruction.}We show the reconstruction results of BTCV with a mask ratio of 75\% in Fig \ref{fig3}.
Notably, the ultimate goal of selfMedHPM is to benefit downstream segmentation tasks rather than generating high-quality reconstructions. \\
\textbf{Multi-organ Segmentation.} The results of abdomen multi-organ segmentation are shown in Table \ref{tab1}. HPM self pre-training improves upon the baseline\cite{3} from 83.5\% to 85.8\% on average DSC. 
The results on SMWB are listed in Table \ref{tab2}. HPM self pre-training improves upon the baseline from 88.5\% to 90.9\% on average DSC. 
At the same time, the number of parameters in our method is less than 1.1 times that of baseline, which is negligible.
Qualitative multi-organ segmentation comparisons are presented in Fig \ref{fig4}. \\
\textbf{Ablation Study.}We evaluated downstream segmentation task performance on SMWB, as shown in Table \ref{tab4}. Note that only predicting loss $ L_{pred}$ as an additional target can improve performance on segmentation tasks, verifying the validity of using the model as a teacher rather than a student.
\begin{table}[htbp]
	\centering
	\caption{The results of ablation experiments on SMWB.}
	\label{tab4}
	\begin{tabular*}{\linewidth}{@{\extracolsep{\fill}}ccc}
	\toprule 
		components & DSC & HD95  \\ 
		\midrule 
		w/o $ L_{pred}$,learn to mask      & 0.885 &3.53  \\ 
		w/o learn to mask      &  0.892 & 3.05   \\
		Ours      & \textbf{0.909} & \textbf{2.65}   \\
	\bottomrule 
	\end{tabular*}
\end{table}
\vspace{-0.1cm}
\section{CONCLUSION}
In this paper, we 
propose selfMedHPM, which introduces an auxiliary reconstruction loss prediction task and thus iteratively guides the training process in a produced and solved manner. In experiments, our method can improve the performance of SOTA on 3D medical CT image segmentation tasks, while taking less than 1.1× time to train against selfMedMAE baseline.
In future work, we will test the efficacy of selfMedHPM in other medical image analysis tasks such as prognosis and outcome prediction tasks\cite{14}. Meanwhile, how to design a loss prediction task without an extra auxiliary decoder can be
further studied.


\vfill
\pagebreak

\section{Compliance with ethical standards}
\label{sec:ethics}
This research study was conducted retrospectively using human subject data made available in open access by (Source information). Ethical approval was not required as confirmed by the license attached with the open access data.

\section{Acknowledgments}
\label{sec:acknowledgments}
This work was supported in part by the National Nature Science Foundation of China under Grant 62271246, and Grant U20A20389; and in part by the Natural Science Foundation of Jiangsu Province under Grant BK20221477.


\begin{thebibliography}{00}
\bibitem{1} K. He et al., “Masked autoencoders are scalable
vision learners,” in CVPR, 2022, pp. 16000–16009.
\bibitem{2} A. Dosovitskiy et al., “An image is worth 16x16
words: Transformers for image recognition at scale,”
arXiv preprint arXiv:2010.11929, 2020.
\bibitem{3} L. Zhou, H. Liu, J. Bae, J. He, D. Samaras and P. Prasanna, "Self pre-training with masked autoencoders for medical image analysis", arXiv:2203.05573, 2022.
\bibitem{4} H. Wang, K. Song, J. Fan, Y. Wang, J. Xie and Z. Zhang, "Hard patches mining for masked image modeling", arXiv:2304.05919, 2023.
\bibitem{5} A. Hatamizadeh et al., “Unetr: Transformers for 3d
medical image segmentation,” in WACV, 2022.
\bibitem{6} A. Taleb, W. Loetzsch, N. Danz, J. Severin, T. Gaertner, B. Bergner, and
C. Lippert, “3d self-supervised methods for medical imaging,” Advances in Neural Information Processing Systems, vol. 33, pp. 18158–18172, 2020.
\bibitem{7} A. Taleb, W. Loetzsch, N. Danz, J. Severin, T. Gaertner, B. Bergner, and
C. Lippert, “3d self-supervised methods for medical imaging,” Advances in Neural Information Processing Systems, vol. 33, pp. 18158–18172, 2020.
\bibitem{8} O. Ronneberger et al., “U-net: Convolutional networks for biomedical image segmentation,” in MICCAI.Springer, 2015, pp. 234–241.
\bibitem{9} J. Chen et al., “Transunet: Transformers make
strong encoders for medical image segmentation,” arXiv
preprint arXiv:2102.04306, 2021.
\bibitem{10} J. Schlemper et al., “Attention gated networks: Learning to leverage salient regions in medical images,” Medical image analysis, vol. 53, pp. 197–207, 2019.
\bibitem{11} Z. Cai et al., “Dstunet: Unet with efficient dense
swin transformer pathway for medical image segmentation,” in ISBI, 2022.

\bibitem{12}  A. Paszke et al., “Pytorch: An imperative style,
high-performance deep learning library,” NeurIPS,
2019
\bibitem{13} MONAI Consortium, “MONAI: Medical Open Network for AI,” 3 2020.
\bibitem{14} Joseph Bae et al., “Predicting mechanical ventilation and mortality in covid-19 using radiomics and deep learning on chest radiographs: A multi-institutional study,” Diagnostics, vol. 11, no. 10, pp. 1812, 2021.


\end{thebibliography}
\end{document}